# Active Learning Enabled Low-cost Cell Image Segmentation Using Bounding Box Annotation

Yu Zhu[a], Qiang Yang[a], Li Xu[a,*]

[a] *College of Electrical Engineering, Zhejiang University, Hangzhou 310027, China*

**Abstract**

Cell image segmentation is usually implemented using fully supervised deep learning methods, which heavily rely on extensive annotated training data. Yet, due to the complexity of cell morphology and the requirement for specialized knowledge, pixel-level annotation of cell images has become a highly labor-intensive task. To address the above problems, we propose an active learning framework for cell segmentation using bounding box annotations, which greatly reduces the data annotation cost of cell segmentation algorithms. First, we generate a box-supervised learning method (denoted as YOLO-SAM) by combining the YOLOv8 detector with the Segment Anything Model (SAM), which effectively reduces the complexity of data annotation. Furthermore, it is integrated into an active learning framework that employs the MC DropBlock method to train the segmentation model with fewer box-annotated samples. Extensive experiments demonstrate that our model saves more than ninety percent of data annotation time compared to mask-supervised deep learning methods.

## 1. Introduction

Image segmentation plays a crucial role in various medical image analysis tasks, which can help in diagnosis, treatment planning and scientific research. Among them, cellular instance segmentation is a key challenge, which directly affects the accuracy of cell quantification, pathological analysis, and personalized medicine [1]. Conventional cell segmentation methods, including level sets and watersheds, are inflexible and unautomated [2]. In contrast, deep learning algorithms address these challenges through end-to-end learning, automatically extracting optimal features from cell images, and achieving higher accuracy without manual feature design by researchers [3]. However, many current deep learning methods for cell segmentation rely heavily on precise mask-supervised training, and the accuracy of the model depends on the quality and quantity of annotated data [4]. Annotating cells in microscopic images is more challenging than the objects in natural images because of their complex cell morphology, unclear boundaries, frequent noise interference, and the need for significant expertise [5]. It is therefore important to explore potential ways to reduce the burden of manual annotation.

Cell segmentation aims at obtaining the location and shape of the cells, but the annotation process is very time-consuming whether it is mask-based or polygon-based [6]. Research has shown that annotating an object's bounding box in COCO only requires 8.8% of the time (7s vs. 79.2s) compared to annotating its mask based on polygons [7]. Hence, more and more research is dedicated to exploring efficient methods for leveraging coarse annotations, such as points [8], bounding boxes [9], and scribbles [10], to obtain cell masks [11,12]. Cell images are diverse and complex, with unclear and overlapping cell boundaries, so box annotation is more accurate than point or scribble annotation and more efficient than polygon annotation, making it an ideal weak annotation method for mask acquisition in cell segmentation tasks [13].

Given the high cost of acquiring annotated data, it is imperative to explore how to utilize the annotation data more efficiently. Since the annotation of cells requires the intervention of experts in the relevant field, active learning as a human-in-the-loop framework fits well with this task. Active learning (AL) aims to select the most useful samples from the unlabeled dataset and hand it over to the oracle for annotation, to reduce the cost of annotation as much as possible while still maintaining performance [14,15]. It has been demonstrated that active learning can significantly reduce the number of training samples required, thereby alleviating the workload of experts for segmentation tasks [15–17]. Especially in the field of cell segmentation, by only annotating a small portion of cell masks, the model can achieve performance that is very close to a mask-supervised method [17,18]. However, the annotated data used by the above method is still a pixel-level mask object, which still requires a lot of annotation time.

Therefore, this paper presents a novel method that only requires bounding box annotations for cell objects and uses active learning to further improve the utilization of annotated samples. Specifically, we combine the object detection model YOLOv8 with the Segment Anything Model (SAM) [19] (denoted as YOLO-SAM) to achieve accurate cell segmentation, and then integrate it into an active learning framework to efficiently utilize box-annotated data. Our main contributions can be summarized as follows.

- A box-supervised deep learning method (YOLO-SAM) for cell segmentation is presented that outperforms other mask-supervised methods such as Mask R-CNN on three public medical datasets.
- Combining YOLO-SAM with active learning, a low-cost algorithmic framework for cell segmentation

is proposed, which greatly reduces the data annotation time.
- The impact of Monte-Carlo DropBlock sampling method in active learning on model learning efficiency is investigated, and the cost of data annotation used in this method is analyzed.

## 2. Related Work

Currently, research has shown that instance segmentation masks can be obtained using only bounding box annotations. The recent Box2Seg [20] method utilizes masks generated by GrabCut as pseudo-tags, iterating to refine the segmentation results. BBTP [21] transformed the box tightness prior into latent ground truth using multiple instance learning (MIL) and utilized structural constraints to preserve piece-wise smoothness in predicted masks. Boxinst [22] replaces the mask loss with the projection loss and pairwise loss in CondInst without modifying the segmentation network itself, achieving a significant improvement in segmentation performance. In addition, BoxLevelSet [23] incorporates the classic level set evolution model into deep neural network learning. It leverages both the input image and its deep features to implicitly evolve the level set curve and employs a local consistency module based on pixel affinity kernels to extract local context and spatial relations. However, the methods above are specifically designed for natural objects and rely largely on CNNs that only capture local details. In contrast, this paper proposes a novel box-supervised approach for cell segmentation, which combines CNN with transformer that considers long-distance dependencies.

YOLOv8 is a CNN-based single-stage object detection model, which improves the detection performance and achieves better accuracy by introducing the CSPDarknet53 architecture and using PANet for feature fusion [24]. YOLOv8 offered a range of five scaled versions, namely YOLOv8n (nano), YOLOv8s (small), YOLOv8m (medium), YOLOv8l (large), and YOLOv8x (extra-large). Segment Anything Model (SAM) is a transformer model trained on the extensive SA1B dataset, and designed to segment an object of interest in an image given certain prompts provided by a user [19]. Prompts can take the form of a single point, a bounding box, or text. In this paper, we propose a box-supervised learning method that uses the bounding box output from the YOLOv8 model as a prompt for SAM.

Active learning aims to explore how to obtain maximum performance gains with minimal labeled samples, focusing on selecting the most informative samples from unlabeled datasets. For a given unlabeled data set, the current main query strategies include diversity-based methods [25–27] and uncertainty-based methods [27–29]. Diversity-based methods select data samples that represent the overall distribution of the data pool. This method is affected by the density of the data pool, has limited ability to improve decision boundaries, and has a relatively high computational cost [30]. Uncertainty-based methods select the N samples with the highest uncertainty by measuring the uncertainty of a pool of unlabeled data [14]. Uncertainty can be determined by computing the entropy of the probability of a data sample [31], the margin of the first and second predicted probability [32], or by using Bayesian deep neural networks with Monte-Carlo (MC) dropout [33–35]. However, many deep learning models (e.g., YOLOv8) are based on convolutional operations. Dropout was found to be ineffective for convolutional neural networks, so DropBlock for convolutional neural networks was proposed, which achieves uncertainty modeling by dropping neurons in contiguous regions of the feature map [36,37]. Hence, we propose Monte-Carlo (MC) DropBlock as an uncertainty sampling method in active learning.

## 3. The Proposed Cell Segmentation Method Based on Active Learning

### 3.1 Overview

As shown in Fig.1, our method is a human-in-the-loop framework that incorporates box supervision into active learning for segmentation tasks. Firstly, A bounding box supervised segmentation algorithm, i.e. YOLO-SAM, is proposed to obtain an accurate mask through the bounding box of a cell. Then, YOLO-SAM is integrated into the active learning framework to observe the effect of different sampling strategies on the efficiency of model performance improvement.

Details of each part of the proposed network structure are described in this section.

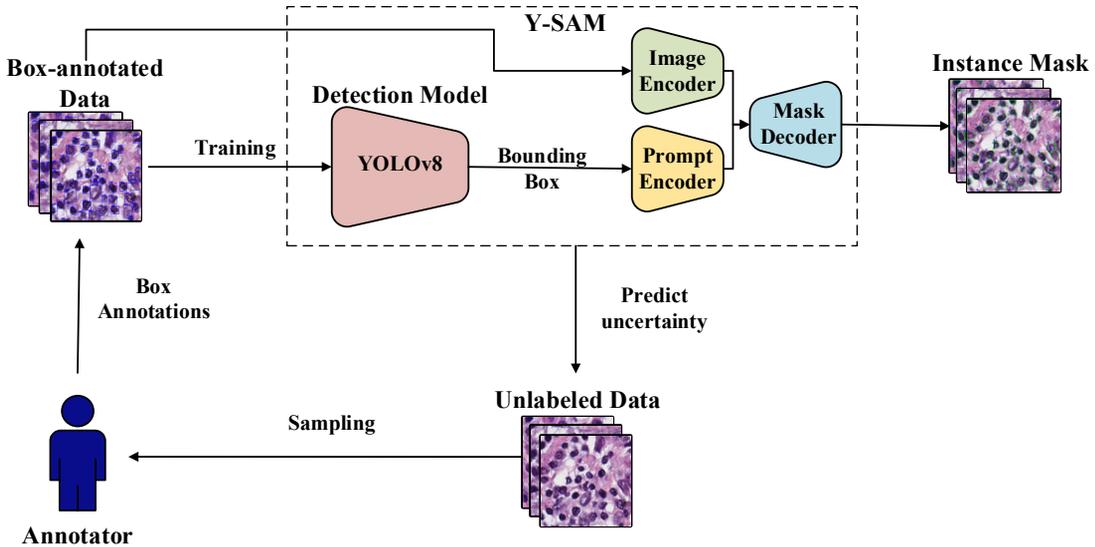

**Fig.1** Overall framework of our method.

*3.2 Box-supervised Segmentation Algorithm YOLO-SAM*

As shown in Fig.1, we trained YOLOv8 for the region of interest detection and used predicted bounding boxes as prompts to SAM for instance segmentation, i.e. YOLO-SAM.

Before training the model, we preprocess the cell images to be used. This includes resizing the images, enhancing the contrast, and removing noise. YOLOv8 utilizes the C2f (cross-level partial bottleneck with two convolutions) module and employs a spatial pyramid fast pooling (SPPF) layer to make the feature pool a fixed-size map. We fed cell images and corresponding bounding box labels to YOLOv8 for iterative training and fine-tuned the weighting parameters for each dataset to optimize cell detection. The main objective of the YOLOv8 model is to accurately detect approximate boundary boxes around the Regions of Interest (ROI) present in our cell images. These boundary boxes serve as essential inputs to the SAM model, performing the subsequent segmentation task.

The reason for aiming at approximate boundary boxes is that the SAM model's dice score remains relatively consistent even when the boundary boxes vary by a small margin, specifically 5-10 pixels [38]. SAM is a versatile and potent architecture designed for real-world segmentation tasks. It consists of the following components: image encoder, mask encoder, prompt encoder and mask decoder [39]. The image encoder converts the input image into encoded features. The mask encoder encodes the mask as a dense prompt, while the prompt encoder encodes the bounding box predicted by YOLOv8 as a sparse prompt. The mask decoder, consisting of multiple layers of attention, interacts with the image features with the prompt features to output the final segmentation map. Utilizing image, prompt, and lightweight mask encoders, SAM accurately predicts segmentation masks. By combining YOLOv8's approximate boundary boxes with the spatial attention mechanisms of SAM, this approach ensures better localization and segmentation of regions of interest in cell images.

*3.3 Active Learning Based Algorithm for Box-supervised Cell Segmentation*

Generally speaking, active learning consists of four main steps in our study. First, we selected an initial set of images $L_0$ from the training pool L, and annotated them by an oracle. Second, we trained an object detector $f(x|L_0)$ using YOLOv8 on these images and predicted segmentation masks by SAM. Third, the trained segmentation model $g(x|L_0)$ was evaluated on the independent test set to determine its performance. Fourth, a new subset $L_i$ of images was selected from the training pool with sampling. After the fourth step, the selected images $L_i$ are annotated and added to the previous training set $L_0$. The segmentation model is retrained on this combined image set $L_0+L_i$ and then evaluated to sample the new images. The whole process is repeated for several iterations, so that there is $\Sigma L_i$ that achieves equivalent performance to L, i.e. $f(x|\Sigma L_i) \approx f(x|L)$. The details of our data augmentation algorithm are presented in Algorithm 1.

---
**Algorithm 1**: Box-supervised Segmentation Method Based on Active Learning
---
Input: data-pool, sample-size, pre-training weight $F_0=0$
Output: DSC-list
1:   initia-data ← RandomlySample(data-pool, sample-size)
2:   $L_0$ ← Annotate(initia-data)
9:   for i ← 1 to loop do
10:     sampled-image ← sampling-strategy (data-pool, sample-size)
11:     $L_i$ ← Annotate(Sampled-image)
12:     $F_i$ ← Train-YOLOv8($L_i+L_{i-1}$, $F_{i-1}$)
13:     Bbox-output ← F(validate-set)
14:     Mask-output ← SAM(Bbox-output)
15:     DSC ← Evaluate-YOLO-SAM(Mask-output)
16:     DSC-list.insert(DSC)
17:     data-pool ← data-pool − sampled-image
18:   end
19:   return DSC-list
---

In this paper, we employ the Monte-Carlo (MC) DropBlock method for deep learning models, which provides the required uncertainty modeling capability when sampling unlabeled images for active learning. The MC DropBlock approach aims to add DropBlock to the model network and apply them during training and inference, which can be interpreted as generating multiple DropBlock architectures for an averaged prediction. Literature [37] derives a proof of theory showing that MC DropBlock is equivalent to a Bayesian convolutional neural network and therefore captures the epistemic uncertainty of out-of-distribution data. We add DropBlock to the YOLOv8 model and analyze the effect of different locations of its addition on the sampling efficiency of active learning. These three locations are at the backbone-neck junction (MC DropBlock 1), during feature fusion of the neck (MC DropBlock 2), and before the prediction of the head (MC DropBlock 3).

We give a visual example of sampling using the MC DropBlock method, as shown in Figure 2. DropBlock drops contiguous regions of the intermediate feature layer with a certain probability, which can be viewed as giving rise to a sub-network of multiple architectures. In the inference stage, we make T forward passes, and each time the resulting subnetwork outputs a different prediction value separately, which we employ to compute the uncertainty of each image. The uncertainty calculations were adopted from Pieter et al. (2022) [40], where the coefficients of the three components—category, bounding box, and mask—are multiplied together, as shown in Eqs. (1-4).

$$c_i = \frac{1}{t}\sum_{i=1}^{t}\left[1 - \frac{-\sum_{i=1}^{m}P(k_i)\cdot \log P(k_i)}{-\sum_{i=1}^{m}\frac{1}{m}\log\frac{1}{m}}\right] \quad (1)$$

$$c_b = \frac{1}{t} \cdot \sum_{i=1}^{t} IoU(\bar{B}(S), B(s_i)) \tag{2}$$

$$c_m = \frac{1}{t} \cdot \sum_{j=1}^{t} IoU(\bar{M}(S), M(s_j)) \tag{3}$$

$$c = c_i \cdot c_b \cdot c_m \tag{4}$$

where $m$ is the number of classes, $t$ is the number of instance sets, $P(k_i)$ is the confidence score of each class. $\bar{B}(S)$ and $\bar{M}(S)$ are the mean bounding box and mask of all instance sets, respectively. $B(s_i)$ and $M(s_j)$ are the corresponding each individual box and mask prediction within that instance set, respectively. In the sampling stage of active learning, we select the N images with the highest uncertainty to be annotated by experts.

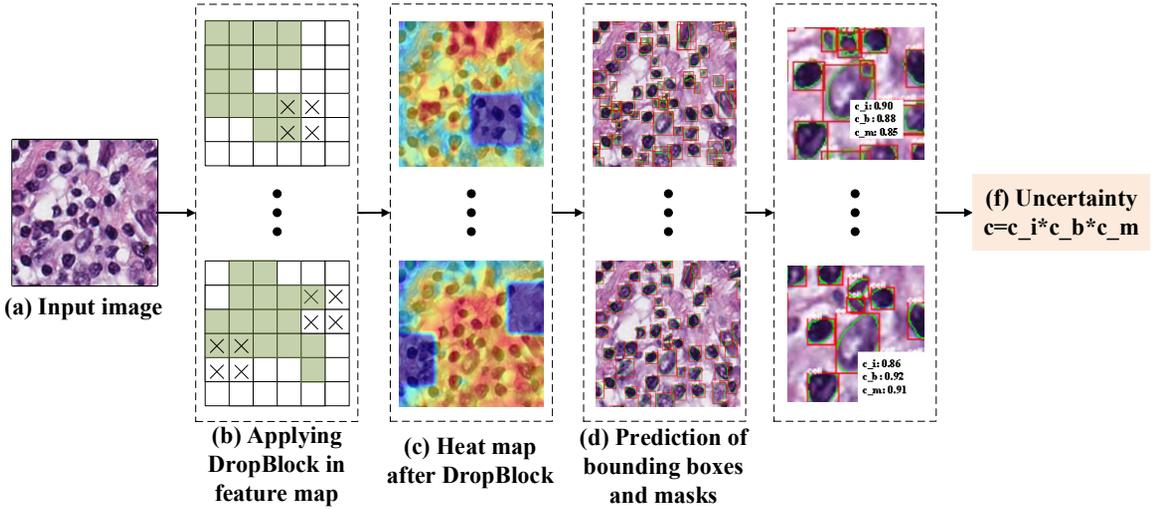

**Fig.2** Visual example of uncertainty calculation with the MC DropBlock method.

## 4. Experiments and Results

*4.1 Datasets and Implementation Details*

To evaluate our proposed method, we conducted extensive experiments on three public cell segmentation datasets, which include PanNuke, Data Science Bowl (DSB) 2018 and MoNuSeg datasets. The details of the datasets are described as follows.

**(1)PanNuke.** PanNuke [41] is a semi automatically generated nuclei instance segmentation and classification dataset with exhaustive nuclei labels across 19 different tissue types. In total, the dataset contains 205,343 labeled nuclei, each with an instance segmentation mask.

**(2)Data Science Bowl (DSB) 2018.** The DSB2018 dataset [42] contains 670 images of cell nuclei with segmentation masks, each containing one nucleus with no overlap between masks. The sample size of the training set and test set are 536 and 134 images, respectively.

**(3)MoNuSeg.** The MoNuSeg dataset [43] contains Multi-Organ Nucleus Segmentation (MoNuSeg) is a nucleus segmentation dataset consisting of H&E images representing nuclei from seven different organs to ensure diversity in nuclear appearance. The dataset consists of a total of 51 images containing 28,846 annotated cells. We segmented the 1000x1000 image into sixteen 250x250 images, resulting in 592 training and validation images and 224 test images.

We used the YOLOv8 medium version as a detector to predict the bounding box, employing a confidence level of 0.2 and an IoU (for NMS) of 0.5, and used ViT-B as the backbone of the SAM-based image encoder. And we compare the proposed YOLO-SAM method with the state-of-the-art box-supervised instance segmentation methods BoxInst [22], Boxlevelset [23] and the classical mask-supervised method Mask R-CNN [44], SOLOv2 [45]. The optimizer for these models uses the AdamW with the learning rate(lr) set to 0.01 at the beginning and gradually decreasing by a factor of 0.01 throughout training with a batch size of 16, and momentum of 0.937, while weight decay is set to 0.0005. In each loop of active learning, YOLO-SAM was trained for 150 epochs. The MC DropBlock method used DropBlock with a dropout probability of 0.25, block size of 7, and a total of 8 forward passes.

We use the Dice Coefficient (DSC) to evaluate the performance of the segmentation model, defined as the similarity between the ground truth mask and the predicted mask. We conduct our experiments using NVIDIA GTX 3090 GPU in Python 3.8.

## 4.2 Results and Analysis

### 4.2.1 Comparison of Box-supervised Learning Approaches

We compared our proposed YOLO-SAM method against the state-of-the-art box-supervised instance segmentation approaches. The results of representative fully mask-supervised methods are also reported for reference.

Table 1 shows the performance comparison of YOLO-SAM with other state-of-the-art methods such as Mask R-CNN and BoxInst on PanNuke, 2018DSB, and MoNuSeg datasets, respectively. On all three datasets, YOLO-SAM outperforms the two representative box-supervised segmentation models, with substantial improvements in segmentation accuracy. Moreover, the performance of YOLO-SAM is almost comparable to the mask-supervised method Mask R-CNN. This is mostly due to the effective combination of a CNN that captures local details and a transformer that considers global information.

Table 1. Performance comparison of different cell segmentation algorithms on PanNuke, 2018DSB, MoNuSeg datasets. The best results are in bold.

|  | Method | PanNuke | 2018DSB | MoNuSeg |
|---|---|---|---|---|
| Mask-supervised methods | Mask R-CNN | 81.02 | 86.26 | 75.75 |
|  | SOLOv2 | 78.25 | 87.80 | 61.54 |
| Box-supervised methods | BoxInst | 69.80 | 77.37 | 62.52 |
|  | Boxlevelset | 75.81 | 83.73 | 76.28 |
|  | Ours(YOLO-SAM) | **80.90** | **88.39** | **78.01** |

Representative visualization results are illustrated in Fig. 3. We can see that segmentation masks predicted by BoxInst has overlapping and undetected cell instances, and segmentation masks generated from BoxLevelset is not clear enough in terms of boundary, but YOLO-SAM can accurately detect cell instances and further improve the shape of segmentation mask. Our method achieves excellent performance.

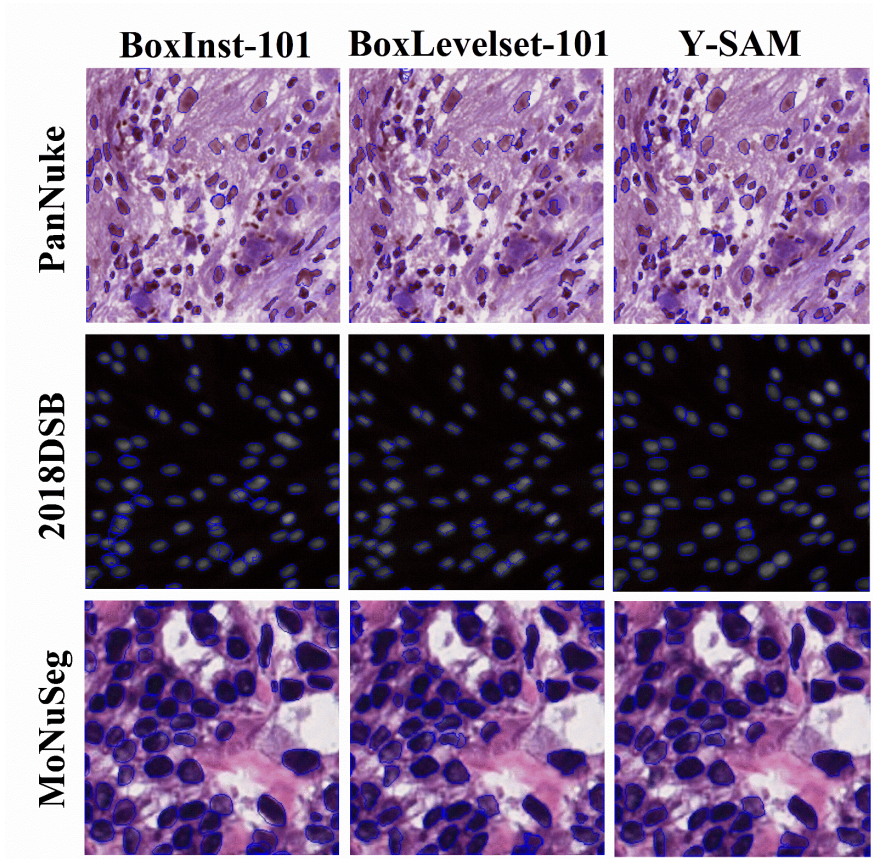

**Fig. 3** Comparison of segmentation masks for random images on PanNuke, 2018DSB and MoNuSeg dataset predicted by BoxInst(ResNet-101), BoxLevelset(ResNet-101), and YOLO-SAM respectively.

#### 4.2.2 Analysis of Annotation Costs Based on Active Learning

We combine the box-supervised deep learning method YOLO-SAM mentioned above with an active learning process to form a weakly supervised learning method. As described in Section 3.3, when employing the MC DropBlock method for sample selection, we investigated the impact of applying DropBlock at three different locations in YOLO-SAM on the efficiency of model performance improvement. The random sampling method is to randomly select the samples to be annotated for training during each round of active learning.

Fig. 4 shows the performance on three public datasets using the MC DropBlock method and the random sampling method, where we counted the DSC values (averaged over three times) as the number of training sets increased. We observed that our method achieves nearly equivalent performance to mask-supervised learning while using minimal data across all three datasets. Notably, the sampling method based on MC DropBlock 1 shows superior performance, as the DropBlock removes semantic information, placing it deeper in the feature extraction network is better for improving generalisation and increasing uncertainty modelling capabilities. On the Pannuke dataset, utilizing only 32.7% of training samples, the sampling method based on MC DropBlock 1 yields a DSC value of 80.1. This highlights that our method achieves 99% of the performance of the mask-supervised instance segmentation model Mask R-CNN while utilizing only 32.7% of the bounding box labeled data. This trend in performance is consistent across the 2018DSB and MoNuSeg datasets as well.

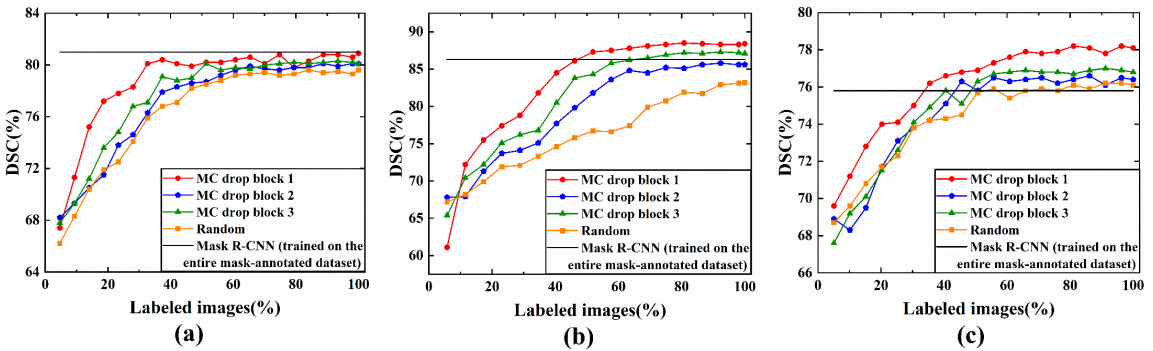

**Fig. 4** Comparison of trends in DSC of our method on (a)PanNuke, (b)2018DSB, and (c)MoNuSeg Datasets. The black solid line represents the performance of the Mask R-CNN model that was trained on the entire mask-annotated training pool.

Literature [7] shows that it takes only 8.8% of the time to annotate an object's bounding box in COCO compared to annotating a mask based on a polygon. As shown in Equation (5), when the DSC of this method reaches 99% of the Mask R-CNN method, we calculated the data annotation time cost at this point compared to the mask-supervised learning method, i.e., Mask R-CNN. The calculated results are shown in Table 2. For example, on the Pannuke dataset, when our model achieves 99% of the performance of Mask R-CNN, the sampling method based on MC DropBlock 1 uses only 32.7% of the box-annotated samples, i.e., it uses only 32.7% *8.8%=2.9% of the annotation time. Thus, compared to mask-supervised segmentation algorithms, our model requires only a few percent of the annotation time to achieve high-performance segmentation, i.e., it saves more than ninety percent of the annotation time.

$$A = \frac{N_b}{N_m} \times 8.8\% \tag{5}$$

where $N_b$ is the number of images annotated with bounding boxes, $N_m$ is the number of images annotated with masks, and 8.8% is the ratio of the average time to annotate the bounding box of a natural object to the time to annotate the polygon mask.

**Table 2.** Minimum annotation time used by each of the sampling methods of active learning on each of the three datasets when the model performance (DSC) reaches 99% of that of Mask-RCNN. The best results are in bold.

| Method \ Datasets | PanNuke | 2018DSB | MoNuSeg |
|---|---|---|---|
| MC DropBlock 1 | **2.9%** | **4.1%** | **2.7%** |
| MC DropBlock 2 | 7.8% | 7.6% | 3.6% |
| MC DropBlock 3 | 4.5% | 5.1% | 3.6% |

The above analysis shows that our method can reduce the number of required training samples, and the annotation time of a single image. Overall, our method greatly reduces the cost of data annotation for cell segmentation.

## 5.   Conclusion and Future Work

This paper presents a method that synergistically combines bounding box annotations with active learning, significantly reducing the annotation cost needed to train a cellular segmentation network. First, a box-supervised segmentation method named YOLO-SAM is designed to achieve accurate cell segmentation using only bounding box annotations. Then, it is integrated into an active learning framework. Applying the MC DropBlock-based active learning approach significantly improves the efficiency of model performance enhancement. Ultimately, we achieve the performance of Mask R-CNN while using only a fraction of the annotation time. This method, which accomplishes cell segmentation with only a small number of box-annotated data, substantially reduces data annotation time cost.

For future work, we will consider several directions in terms of low-cost cell segmentation. First, we will analyze the time cost required for cell segmentation using more active learning sampling strategies. Then, the evaluation method of data annotation cost will be further refined to make the results more scientific.

**References**

[1] F. Li, X. Zhou, J. Ma, S.T.C. Wong, Multiple Nuclei Tracking Using Integer Programming for Quantitative Cancer Cell Cycle Analysis, IEEE Trans. Med. Imaging 29 (2010) 96–105. https://doi.org/10.1109/TMI.2009.2027813.

[2] J. Xu, D. Zhou, D. Deng, J. Li, C. Chen, X. Liao, G. Chen, P.A. Heng, Deep Learning in Cell Image Analysis, Intell. Comput. 2022 (2022) 2022/9861263. https://doi.org/10.34133/2022/9861263.

[3] D. Shen, G. Wu, H.-I. Suk, Deep Learning in Medical Image Analysis, Annu. Rev. Biomed. Eng. 19 (2017) 221–248. https://doi.org/10.1146/annurev-bioeng-071516-044442.

[4] T. Wen, B. Tong, Y. Liu, T. Pan, Y. Du, Y. Chen, S. Zhang, Review of research on the instance segmentation of cell images, Comput. Methods Programs Biomed. 227 (2022) 107211. https://doi.org/10.1016/j.cmpb.2022.107211.

[5] A review of current systems for annotation of cell and tissue images in digital pathology, Biocybern. Biomed. Eng. 41 (2021) 1436–1453. https://doi.org/10.1016/j.bbe.2021.04.012.

[6] W. Du, Y. Huo, R. Zhou, Y. Sun, S. Tang, X. Zhao, Y. Li, G. Li, Consistency label-activated region generating network for weakly supervised medical image segmentation, Comput. Biol. Med. 173 (2024) 108380. https://doi.org/10.1016/j.compbiomed.2024.108380.

[7] T.-Y. Lin, M. Maire, S. Belongie, J. Hays, P. Perona, D. Ramanan, P. Dollár, C.L. Zitnick, Microsoft COCO: Common Objects in Context, in: D. Fleet, T. Pajdla, B. Schiele, T. Tuytelaars (Eds.), Comput. Vis. – ECCV 2014, Springer International Publishing, Cham, 2014: pp. 740–755. https://doi.org/10.1007/978-3-319-10602-1_48.

[8] T. Zhao, Z. Yin, Weakly Supervised Cell Segmentation by Point Annotation, IEEE Trans. Med. Imaging 40 (2021) 2736–2747. https://doi.org/10.1109/TMI.2020.3046292.

[9] J. Wang, B. Xia, Weakly supervised image segmentation beyond tight bounding box annotations, Comput. Biol. Med. 169 (2024) 107913. https://doi.org/10.1016/j.compbiomed.2023.107913.

[10] J. Ying, W. Huang, L. Fu, H. Yang, J. Cheng, Weakly supervised segmentation of uterus by scribble labeling on endometrial cancer MR images, Comput. Biol. Med. 167 (2023) 107582. https://doi.org/10.1016/j.compbiomed.2023.107582.

[11] P. Shrestha, N. Kuang, J. Yu, Efficient end-to-end learning for cell segmentation with machine generated weak annotations, Commun. Biol. 6 (2023) 1–10. https://doi.org/10.1038/s42003-023-04608-5.

[12] S.M. Ha, Y.S. Ko, Y. Park, Generating BlobCell Label from Weak Annotations for Precise Cell Segmentation, in: S.-A. Ahmadi, S. Pereira (Eds.), Graphs Biomed. Image Anal. Overlapped Cell Tissue Dataset Histopathol., Springer Nature Switzerland, Cham, 2024: pp. 161–170. https://doi.org/10.1007/978-3-031-55088-1_15.

[13] L. Yang, Y. Zhang, Z. Zhao, H. Zheng, P. Liang, M.T.C. Ying, A.T. Ahuja, D.Z. Chen, BoxNet: Deep Learning Based Biomedical Image Segmentation Using Boxes Only Annotation, (2018). https://doi.org/10.48550/arXiv.1806.00593.

[14] P. Ren, Y. Xiao, X. Chang, P.-Y. Huang, Z. Li, B.B. Gupta, X. Chen, X. Wang, A Survey of Deep Active Learning, ACM Comput. Surv. 54 (2021) 180:1-180:40. https://doi.org/10.1145/3472291.

[15] A. Biswas, N.M. Abdullah Al, M.S. Ali, I. Hossain, M.A. Ullah, S. Talukder, Active Learning on Medical Image, in: B. Zheng, S. Andrei, M.K. Sarker, K.D. Gupta (Eds.), Data Driven Approaches Med. Imaging, Springer Nature Switzerland, Cham, 2023: pp. 51–67. https://doi.org/10.1007/978-3-031-47772-0_3.

[16] M. Gaillochet, C. Desrosiers, H. Lombaert, Active learning for medical image segmentation with stochastic batches, Med. Image Anal. 90 (2023) 102958. https://doi.org/10.1016/j.media.2023.102958.

[17] A. Chowdhury, S.K. Biswas, S. Bianco, Active deep learning reduces annotation burden in automatic cell segmentation, in: Med. Imaging 2021 Digit. Pathol., SPIE, 2021: pp. 94–99. https://doi.org/10.1117/12.2579537.

[18] T. Kim, K.H. Lee, S. Ham, B. Park, S. Lee, D. Hong, G.B. Kim, Y.S. Kyung, C.-S. Kim, N. Kim, Active learning for accuracy enhancement of semantic segmentation with CNN-corrected label curations: Evaluation on kidney segmentation in abdominal CT, Sci. Rep.

10 (2020) 366. https://doi.org/10.1038/s41598-019-57242-9.

[19] A. Kirillov, E. Mintun, N. Ravi, H. Mao, C. Rolland, L. Gustafson, T. Xiao, S. Whitehead, A.C. Berg, W.-Y. Lo, P. Dollar, R. Girshick, Segment Anything, in: 2023: pp. 4015–4026. https://openaccess.thecvf.com/content/ICCV2023/html/Kirillov_Segment_Anything_ICCV_2023_paper.html (accessed March 12, 2024).

[20] V. Kulharia, S. Chandra, A. Agrawal, P. Torr, A. Tyagi, Box2Seg: Attention Weighted Loss and Discriminative Feature Learning for Weakly Supervised Segmentation, in: A. Vedaldi, H. Bischof, T. Brox, J.-M. Frahm (Eds.), Comput. Vis. – ECCV 2020, Springer International Publishing, Cham, 2020: pp. 290–308. https://doi.org/10.1007/978-3-030-58583-9_18.

[21] C.-C. Hsu, K.-J. Hsu, C.-C. Tsai, Y.-Y. Lin, Y.-Y. Chuang, Weakly Supervised Instance Segmentation using the Bounding Box Tightness Prior, in: Adv. Neural Inf. Process. Syst., Curran Associates, Inc., 2019. https://proceedings.neurips.cc/paper_files/paper/2019/hash/e6e713296627dff6475085cc6a224464-Abstract.html (accessed March 9, 2024).

[22] Z. Tian, C. Shen, X. Wang, H. Chen, BoxInst: High-Performance Instance Segmentation With Box Annotations, in: 2021: pp. 5443–5452. https://openaccess.thecvf.com/content/CVPR2021/html/Tian_BoxInst_High-Performance_Instance_Segmentation_With_Box_Annotations_CVPR_2021_paper.html (accessed December 28, 2023).

[23] W. Li, W. Liu, J. Zhu, M. Cui, R. Yu, X. Hua, L. Zhang, Box2Mask: Box-supervised Instance Segmentation via Level-set Evolution, (2022). https://doi.org/10.48550/arXiv.2212.01579.

[24] J. Terven, D. Cordova-Esparza, A Comprehensive Review of YOLO Architectures in Computer Vision: From YOLOv1 to YOLOv8 and YOLO-NAS, Mach. Learn. Knowl. Extr. 5 (2023) 1680–1716. https://doi.org/10.3390/make5040083.

[25] Z. Liang, X. Xu, S. Deng, L. Cai, T. Jiang, K. Jia, Exploring Diversity-based Active Learning for 3D Object Detection in Autonomous Driving, (2022). https://doi.org/10.48550/arXiv.2205.07708.

[26] Q. Jin, M. Yuan, Q. Qiao, Z. Song, One-shot active learning for image segmentation via contrastive learning and diversity-based sampling, Knowl.-Based Syst. 241 (2022) 108278. https://doi.org/10.1016/j.knosys.2022.108278.

[27] U. Patel, H. Dave, V. Patel, Hyperspectral Image Classification using Uncertainty and Diversity based Active Learning, Scalable Comput. Pract. Exp. 22 (2021) 283–293. https://doi.org/10.12694/scpe.v22i3.1865.

[28] S. Ma, H. Wu, A. Lawlor, R. Dong, Breaking the Barrier: Selective Uncertainty-based Active Learning for Medical Image Segmentation, (2024). https://doi.org/10.48550/arXiv.2401.16298.

[29] S. Hwang, J. Choi, J. Choi, Uncertainty-Based Selective Clustering for Active Learning, IEEE Access 10 (2022) 110983–110991. https://doi.org/10.1109/ACCESS.2022.3216065.

[30] O. Sener, S. Savarese, Active Learning for Convolutional Neural Networks: A Core-Set Approach, (2018). https://doi.org/10.48550/arXiv.1708.00489.

[31] Y. Siddiqui, J. Valentin, M. Niessner, ViewAL: Active Learning With Viewpoint Entropy for Semantic Segmentation, in: 2020: pp. 9433–9443. https://openaccess.thecvf.com/content_CVPR_2020/html/Siddiqui_ViewAL_Active_Learning_With_Viewpoint_Entropy_for_Semantic_Segmentation_CVPR_2020_paper.html (accessed March 9, 2024).

[32] B. Miller, F. Linder, W.R. Mebane, Active Learning Approaches for Labeling Text: Review and Assessment of the Performance of Active Learning Approaches, Polit. Anal. 28 (2020) 532–551. https://doi.org/10.1017/pan.2020.4.

[33] S. Schmidt, Q. Rao, J. Tatsch, A. Knoll, Advanced Active Learning Strategies for Object Detection, in: 2020 IEEE Intell. Veh. Symp. IV, 2020: pp. 871–876. https://doi.org/10.1109/IV47402.2020.9304565.

[34] A. Kendall, Y. Gal, What Uncertainties Do We Need in Bayesian Deep Learning for Computer Vision?, (2017). https://doi.org/10.48550/arXiv.1703.04977.

[35] S. Belharbi, I. Ben Ayed, L. McCaffrey, E. Granger, Deep Active Learning for Joint Classification & Segmentation With Weak Annotator, in: 2021: pp. 3338–3347.

https://openaccess.thecvf.com/content/WACV2021/html/Belharbi_Deep_Active_Learning_for_Joint_Classification__Segmentation_With_Weak_WACV_2021_paper.html (accessed March 9, 2024).

[36] G. Ghiasi, T.-Y. Lin, Q.V. Le, DropBlock: A regularization method for convolutional networks, (2018). https://doi.org/10.48550/arXiv.1810.12890.

[37] K. Deepshikha, S.H. Yelleni, P.K. Srijith, C.K. Mohan, Monte Carlo DropBlock for Modelling Uncertainty in Object Detection, (2021). https://doi.org/10.48550/arXiv.2108.03614.

[38] S. Pandey, K.-F. Chen, E.B. Dam, Comprehensive Multimodal Segmentation in Medical Imaging: Combining YOLOv8 with SAM and HQ-SAM Models, (2023). https://doi.org/10.48550/arXiv.2310.12995.

[39] A. Kirillov, E. Mintun, N. Ravi, H. Mao, C. Rolland, L. Gustafson, T. Xiao, S. Whitehead, A.C. Berg, W.-Y. Lo, P. Dollár, R. Girshick, Segment Anything, (2023). https://doi.org/10.48550/arXiv.2304.02643.

[40] P.M. Blok, G. Kootstra, H.E. Elghor, B. Diallo, F.K. van Evert, E.J. van Henten, Active learning with MaskAL reduces annotation effort for training Mask R-CNN, Comput. Electron. Agric. 197 (2022) 106917. https://doi.org/10.1016/j.compag.2022.106917.

[41] J. Gamper, N.A. Koohbanani, K. Benes, S. Graham, M. Jahanifar, S.A. Khurram, A. Azam, K. Hewitt, N. Rajpoot, PanNuke Dataset Extension, Insights and Baselines, (2020). https://doi.org/10.48550/arXiv.2003.10778.

[42] J.C. Caicedo, A. Goodman, K.W. Karhohs, B.A. Cimini, J. Ackerman, M. Haghighi, C. Heng, T. Becker, M. Doan, C. McQuin, M. Rohban, S. Singh, A.E. Carpenter, Nucleus segmentation across imaging experiments: the 2018 Data Science Bowl, Nat. Methods 16 (2019) 1247–1253. https://doi.org/10.1038/s41592-019-0612-7.

[43] N. Kumar, R. Verma, D. Anand, Y. Zhou et al., A Multi-Organ Nucleus Segmentation Challenge, IEEE Trans. Med. Imaging 39 (2020) 1380–1391. https://doi.org/10.1109/TMI.2019.2947628.

[44] K. He, G. Gkioxari, P. Dollar, R. Girshick, Mask R-CNN, in: 2017: pp. 2961–2969. https://openaccess.thecvf.com/content_iccv_2017/html/He_Mask_R-CNN_ICCV_2017_paper.html (accessed March 5, 2024).

[45] X. Wang, R. Zhang, T. Kong, L. Li, C. Shen, SOLOv2: Dynamic and Fast Instance Segmentation, in: Adv. Neural Inf. Process. Syst., Curran Associates, Inc., 2020: pp. 17721–17732. https://proceedings.neurips.cc/paper/2020/hash/cd3afef9b8b89558cd56638c3631868a-Abstract.html (accessed March 5, 2024).